\documentclass{article}
\usepackage{multirow}
\usepackage{graphicx}
\usepackage{wrapfig}
\usepackage{iclr2027_conference,times}

\usepackage{amsmath,amsfonts,bm}

\def\eqref#1{equation~\ref{#1}}

\def\1{\bm{1}}

\DeclareMathAlphabet{\mathsfit}{\encodingdefault}{\sfdefault}{m}{sl}
\SetMathAlphabet{\mathsfit}{bold}{\encodingdefault}{\sfdefault}{bx}{n}

\usepackage[hidelinks]{hyperref}
\usepackage{url}
\usepackage{booktabs}

\title{Text-Assisted Event Stream Visual Place Recognition via Multi-Scale Spatial-Temporal Optimal Transport}

\author{%
\parbox{0.94\textwidth}{%
\raggedright\normalfont
\textbf{Lan Chen}\textsuperscript{1}\quad
\textbf{Xingxing Xiong}\textsuperscript{2}\quad
\textbf{Xiao Wang}\textsuperscript{2,*}\quad
\textbf{Ziyan Zhang}\textsuperscript{2,*}\\
\textbf{Zikang Yan}\textsuperscript{2}\quad
\textbf{Yuanhao Zheng}\textsuperscript{2}\quad
\textbf{Bo Jiang}\textsuperscript{2}\quad
\textbf{Yonghong Tian}\textsuperscript{3}\\[5pt]
{\small
\textsuperscript{1}School of Electronic and Information Engineering,
Anhui University, Hefei, China\\
\textsuperscript{2}School of Computer Science and Technology,
Anhui University, Hefei, China\\
\textsuperscript{3}School of Computer Science,
Peking University, Beijing, China\\[4pt]}
{\footnotesize\ttfamily
\{chenlan,wangxiao,jiangbo\}@ahu.edu.cn,\\
\{e24201089,e126112022,e126221264\}@stu.ahu.edu.cn,\\
zhangziyanahu@163.com,\quad yhtian@pku.edu.cn\\[3pt]}
}}

\iclrfinalcopy 
\begin{document}

\maketitle

\begin{abstract}
Benefiting from the high dynamic range, high temporal resolution, and low energy consumption of event cameras, low‑altitude economic research leveraging event cameras has become a popular topic. Several existing attempts introduce natural‑language descriptions into visual place recognition (VPR) to mitigate the sparsity of structural spatial information inherent to event cameras. Nevertheless, these methods still fail to fully address two core challenges: effectively encoding event streams and semantically aligning event streams with text. In this paper, we propose a multi‑scale spatial‑temporal optimal transport approach, MSTOT‑EPR, to address these challenges. Specifically, given event streams and scene text information, we first adopt a visual encoder and a text encoder to extract respective feature representations. We then apply optimal-transport constraints to achieve cross‑modal alignment. We then design a cross‑modal bottleneck fusion module to further fuse bimodal features. Furthermore, we develop multi‑scale spatial optimal transport and inter‑frame temporal optimal transport modules to mine high‑quality spatial‑temporal event features for scene recognition. Extensive experiments on the NYC‑Event‑VPR and EPRBench datasets fully validate the effectiveness of our proposed framework for event‑based VPR. 
The source code of this paper will be released upon acceptance.
\end{abstract}

\section{Introduction}

Visual Place Recognition (VPR)~\cite{chen2026sage} aims to match a query image to its corresponding location in a large‑scale geotagged database. It holds great research value for many real‑world applications, including autonomous driving navigation, large‑scale map reconstruction, and low‑altitude economy. State‑of‑the‑art VPR solutions are mostly implemented with visible‑light cameras; thus, they inherit various inherent challenges from conventional visual perception, such as extreme viewpoint variation, illumination fluctuation, adverse weather conditions, long‑term temporal shift, and frequent dynamic occlusions. Developing high‑performance VPR approaches remains a challenging problem worthy of further investigation.

Benefiting from the unique advantages of event cameras, including low energy consumption, high dynamic range, and high temporal resolution, event-based vision has attracted increasing attention from researchers~\cite{gallego2020eventsurvey, chen2026DGVHeat, wang2026rgbePAR}. Each pixel of an event camera works independently and records the intensity changes of scenes continuously. Specifically, an individual event is formulated as $(x, y, t, p)$, where $x, y$ denote the spatial pixel coordinates, $t$ represents the timestamp, and $p$ refers to the event polarity. The polarity value of +1 indicates an increase in brightness, while -1 corresponds to a decrease in brightness. In recent years, several studies have begun to explore event-based VPR~\cite{wang2026eprbench, event-vpr}, such as Sparse-Event-VPR~\cite{FischerRAL2022ICRA2023}, which performs training-free place matching using sparse event-count representations, and EventGeM~\cite{hines2026eventgem}, which combines regional GeM-based global retrieval with local feature matching and RANSAC-based homography re-ranking.

Despite this progress, these works are still limited by the following issues: 
1). Existing event stream-based visual place recognition algorithms mostly adopt event representations from a single time window and rarely consider the asynchronous nature of event cameras. Consequently, they are vulnerable to motion-induced positional shifts and instabilities stemming from event sparsity. 
2). Few existing works take into account the lack of semantic information caused by the spatial sparsity of event streams, while the visual place recognition task heavily relies on both spatial‑detail features and semantic features. 
3). Existing text-enhanced methods for event streams~\cite{} adopt coarse-grained modeling strategies, which fail to establish fine alignment between the temporal features of events and the deep semantic information of texts. These methods cannot effectively adapt to the semantic modeling of complex event streams, thereby severely limiting the performance upper bound of event stream understanding and reasoning tasks.


\begin{figure}[t]
    \centering
    \includegraphics[width=0.88\linewidth]{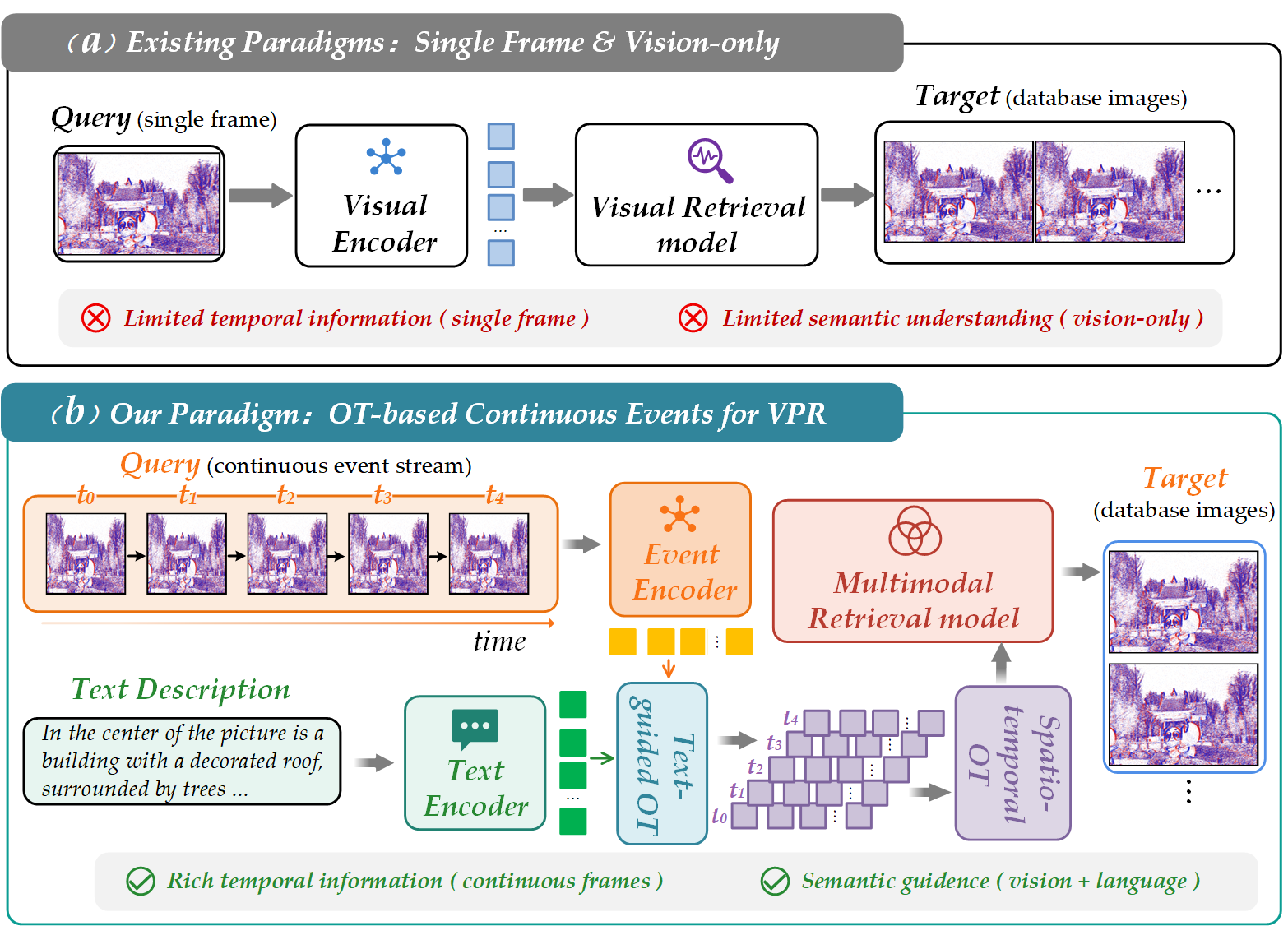}
    \caption{Comparison between conventional vision-only VPR and our event-text VPR paradigm.}
    \label{fig:placeholder}
\end{figure}

To handle these aforementioned issues, in this paper, we propose a novel multi‑scale spatial-temporal optimal transport (OT) based framework for event stream-based VPR, termed MSTOT‑EPR. Optimal transport (OT) was originally proposed to find the minimum‑cost mapping between two probability distributions, resolving the problem of measuring discrepancy for heterogeneous mass distributions. For spatial-temporal modeling, OT excels at capturing structural correspondences across sequential states rather than simple pointwise differences, making it robust against misalignment and spatial-temporal shift. 
Specifically, given the event streams and scene text information, we first adopt a vision encoder and a text encoder to extract their feature representations. Then, we apply optimal transport constraints to achieve cross‑modal alignment. More importantly, we design the cross‑modal bottleneck fusion module to achieve bimodal feature fusion and develop multi‑scale spatial optimal transport and inter‑frame temporal optimal transport modules to mine high‑quality spatial‑temporal event features for scene recognition. 
An overview of our framework can be found in Fig.~\ref{fig:framework}.

To sum up, the main contributions of this paper can be summarized as follows: 

1). We propose a novel multi-scale spatial-temporal optimal transport-based framework for event stream-based visual place recognition, termed MSTOT-EPR. The spatial multi-scale OT and temporal inter-frame OT modules are designed to construct feature-and-position-aware cross-scale transport plans and anchor-guided token-level transport plans with gated temporal aggregation, respectively.

2). We introduce the scenario text as an auxiliary modality and further propose a new optimal transport-based cross-modal alignment module to augment the semantic information of event streams. 

3). Extensive experiments on multiple event-based VPR benchmark datasets fully validated the effectiveness of our proposed framework for the representation learning of text-assisted event streams. 

\section{Related Work} \label{sec::relatedWorks}

\textbf{Visual Place Recognition.~} 
Visual Place Recognition (VPR) is a fundamental capability for autonomous navigation and spatial AI, aiming to identify the geographical location of a query observation by retrieving matching candidates from a reference database. Early learning-based approaches mainly focused on learning compact and discriminative global descriptors from RGB images. NetVLAD~\cite{arandjelovic2016netvlad} introduced a differentiable VLAD layer for weakly supervised place recognition, while subsequent methods such as ConvAP~\cite{ali2022convap} and MixVPR~\cite{ismail2022mixvpr} further improved feature aggregation and descriptor robustness. With the development of transformer architectures, ViT-based models such as TransVPR~\cite{wang2022transvpr} have enhanced long-range dependency modeling and feature alignment for challenging scenes. Recently, foundation models and query-based aggregation strategies have further advanced VPR. AnyLoc~\cite{keetha2023anyloc} demonstrates that off-the-shelf self-supervised visual features can provide strong zero-shot place recognition performance. BoQ~\cite{Ali-bey_2024_CVPR} introduces a bag-of-queries mechanism to aggregate patch-level features into compact global descriptors softly. SALAD~\cite{mahjoub2023salad} explores optimal-transport-inspired aggregation for dynamic feature clustering, improving robustness under challenging appearance variations.

Despite these advances, most existing VPR methods are designed for synchronous RGB frames and rely heavily on dense appearance cues. Their performance can degrade under low illumination, high dynamic range, high-speed motion, and severe motion blur. Moreover, existing aggregation strategies mainly operate on spatial image patches, leaving the sparse and asynchronous temporal structure of event streams insufficiently explored. In contrast, our work focuses on event-stream-based VPR and introduces a spatial-temporal alignment framework tailored to the unique characteristics of event data.

\textbf{Event-based and Multimodal Perception.~}    
Event cameras are bio-inspired asynchronous sensors that respond to per-pixel log-intensity changes, providing microsecond-level temporal resolution, high dynamic range, low latency, and resilience to motion blur~\cite{gallego2020event}. These properties make event streams attractive for robust visual perception in extreme scenarios where conventional RGB cameras may fail. Early event-based methods usually converted raw events into frame-like representations, such as event histograms, time surfaces, or voxel grids, enabling the use of conventional convolutional architectures~\cite{rebecq2019events}. Later studies explored graph-based, point-based, and spiking neural models to better preserve the sparse and asynchronous nature of event data~\cite{bi2019graph}. In the context of place recognition, EventVLAD~\cite{fischer2020eventvlad} and recent event-based VPR benchmarks~\cite{lee2023event} have shown the potential of event cameras for robust localization under challenging visual conditions. However, event streams mainly encode brightness changes rather than absolute appearance, making them semantically sparse and difficult to interpret compared with RGB images. To compensate for limited semantic information, multimodal learning has become an effective strategy. Vision-language pre-training models such as CLIP~\cite{radford2021learning} map visual and textual information into a shared representation space, enabling strong cross-modal generalization. Text-guided and language-assisted recognition methods further demonstrate that natural language descriptions can provide high-level semantic priors for visual understanding~\cite{zhou2022learning,liang2022mind}. In VPR, language descriptions can help disambiguate visually similar places by introducing scene-level contextual cues.

Nevertheless, directly applying vision-language fusion to event-based VPR remains challenging. Event features are sparse, asynchronous, and temporally ordered, whereas textual features are discrete and semantic. Simple concatenation, projection, or contrastive matching may ignore the fine-grained spatial correspondence and temporal evolution of event streams. Therefore, an effective text-assisted event VPR framework should explicitly align multimodal representations across both spatial and temporal dimensions. This motivates our multi-scale spatial-temporal optimal transport design.

\textbf{Optimal Transport in Computer Vision.~} \label{sec:ot_related}
Optimal transport (OT) establishes soft correspondences between probability distributions by minimizing the cost of transporting mass under marginal constraints~\cite{villani2009optimal,peyre2019computational}. Unlike pointwise similarity or global pooling, OT captures distribution-level structure and
supports many-to-many matching, making it suitable for aligning heterogeneous visual representations. Entropic regularization further enables efficient and
differentiable optimization~\cite{cuturi2013sinkhorn}, facilitating applications in domain adaptation and structured feature matching~\cite{courty2017optimal,sarlin2020superglue}. Existing OT-based VPR methods, such as SALAD~\cite{mahjoub2023salad}, primarily focus on spatial aggregation from static RGB images. In contrast, our approach extends OT to event-text
alignment, multi-scale spatial correspondence, and inter-frame temporal modeling. Further background and a detailed discussion of OT-related vision methods are provided in Appendix~\ref{app:ot_background}.

\section{Methodology}  \label{sec::method}

\subsection{Overview}  

\begin{figure}[t]   
\centering
\includegraphics[width=1\linewidth]{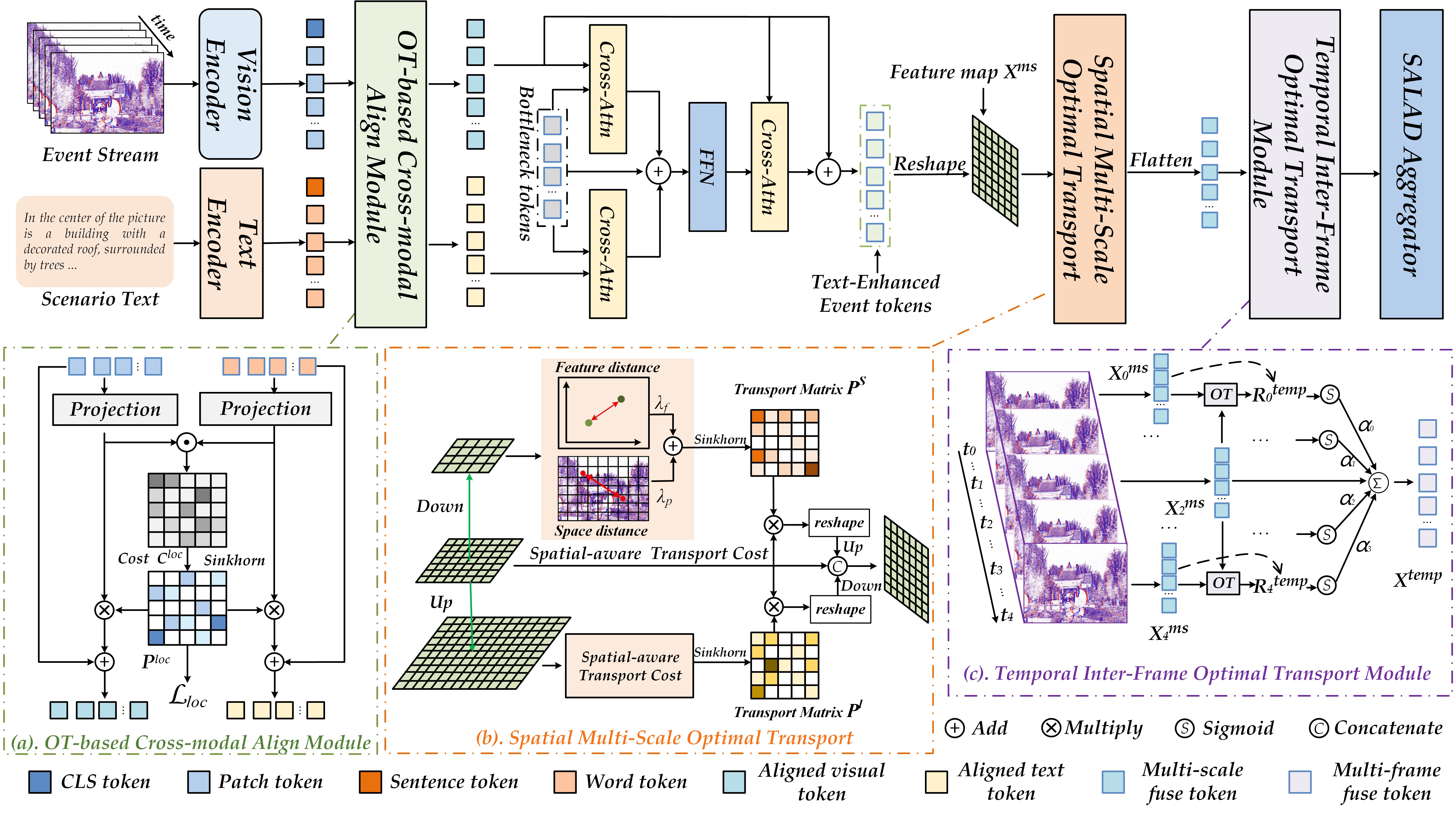}
\caption{An overview of our proposed text-assisted event stream visual place recognition via multi-scale
spatial-temporal optimal transport, termed MSTOT-EPR. The framework integrates OT-based cross-modal alignment with bottleneck fusion, multi-scale spatial-aware OT, and temporal inter-frame OT, followed by SALAD-based feature aggregation.} 
\label{fig:framework}
\end{figure}

Event streams are robust to illumination changes and rapid motion, but their spatial sparsity limits the extraction of discriminative place representations. Textual descriptions provide complementary scene semantics, yet the modality gap between event and text features, together with the temporal variations across event frames, remains challenging to model. To address these issues, we propose Multi-Scale spatial-temporal Optimal Transport for Event-stream Place Recognition (MSTOT-EPR), which establishes structured correspondences across modalities, spatial scales, and time using optimal transport. As illustrated in Fig.~\ref{fig:framework}, we first extract event and text features and align them through cross-modal OT and bottleneck fusion. Multi-scale spatial OT then captures spatial structures at different resolutions, while inter-frame temporal OT aligns consecutive frames to a shared anchor support. Finally, the enhanced spatial-temporal features are aggregated into a compact global descriptor for place retrieval.

\subsection{Input Representations}
\label{sec:input_representations}
\noindent\textbf{Visual representation.}
Given an event stream $\mathcal{E}=\{(x_i,y_i,t_i,p_i)\}_{i=1}^{N}$, where $(x_i,y_i)$, $t_i$, and $p_i\in\{-1,+1\}$ denote the spatial location, timestamp, and polarity of the $i$-th event, respectively, we partition it into $T$ temporal segments and construct an event-frame sequence $\{\mathbf{I}_t\}_{t=0}^{T}$. For each frame $\mathbf{I}_t\in\mathbb{R}^{H\times W}$, the DINOv2 visual encoder~\cite{oquab2023dinov2} extracts a global CLS token $\mathbf{c}_t\in\mathbb{R}^{C_v}$ and a set of patch tokens $\mathbf{V}_t=[\mathbf{v}_{t,1},\ldots,\mathbf{v}_{t,M}]\in\mathbb{R}^{M\times C_v}$, where $M=HW/p^2$ and $p$ is the patch size. The CLS token summarizes frame-level content, while the patch tokens preserve local spatial structure.

\noindent\textbf{Textual representation.}
For the paired scene description $\mathcal{T}$, the CLIP text encoder~\cite{radford2021learning} produces a sentence embedding $\mathbf{s}\in\mathbb{R}^{C_s}$, word-token features $\mathbf{W}=[\mathbf{w}_1,\ldots,\mathbf{w}_L]\in\mathbb{R}^{L\times C_w}$, and an attention mask $\mathbf{m}\in\{0,1\}^{L}$. We discard special and padded tokens according to $\mathbf{m}$, retaining $L'$ valid word tokens. The sentence embedding provides global semantic supervision, whereas the valid word tokens support fine-grained cross-modal alignment.

\subsection{OT-Aligned Modality Fusion}
\label{sec:ot_aligned_fusion}
To bridge the representation gap between event and text modalities,
we first establish fine-grained cross-modal correspondences via optimal transport and then integrate the aligned features through
bottleneck-based fusion.
Specifically, 
we first project the visual patch tokens $\mathbf{V}_t$ and word tokens $\mathbf{W}$ into a shared $D$-dimensional space using two learnable projections $g_v$ and $g_w$.
Their $\ell_2$-normalized representations are denoted by $\hat{\mathbf{v}}_{t,i}=\mathrm{Norm}(g_v(\mathbf{v}_{t,i}))$ and $\hat{\mathbf{w}}_j=\mathrm{Norm}(g_w(\mathbf{w}_j))$, respectively.
To quantify the local discrepancy between the $i$-th visual patch of frame $t$ and the $j$-th valid word token, we define their
pairwise matching cost using cosine distance:
\begin{equation}
    \mathbf{C}^{loc}_{t,ij}=1-\hat{\mathbf{v}}_{t,i}^{\top}\hat{\mathbf{w}}_j.
    \label{eq:local_ot_cost}
\end{equation}
where the pairwise costs are collected into the cost matrix $\mathbf{C}^{\mathrm{loc}}_t$.
Based on the cost matrix, we formulate the fine-grained cross-modal alignment as the following entropically
regularized optimal transport problem:
\begin{equation}
\small{
    \mathbf{P}^{loc}_t
    =\arg\min_{\mathbf{P}\in\mathcal{U}(\boldsymbol{\mu},\boldsymbol{\nu})}
    \langle\mathbf{P},\mathbf{C}^{loc}_t\rangle
    +\varepsilon\sum_{i,j}{P}_{ij}
    (\log{P}_{ij}-1),}
    \label{eq:local_ot_plan}
\end{equation}
where $\boldsymbol{\mu}$ and $\boldsymbol{\nu}$ assign uniform mass to the visual patches and valid word tokens, respectively, i.e., $\mu_i=1/M$ and $\nu_j=1/L'$. 
$\varepsilon>0$ controls the strength of the entropic regularization.

We solve~\eqref{eq:local_ot_plan} using the numerically
stable log-domain Sinkhorn algorithm
\citep{cuturi2013sinkhorn,schmitzer2019stabilized}.
The detailed iterative updates are provided in
Appendix~\ref{app:sinkhorn}.

For brevity, we denote the resulting transport plan as
\begin{equation}
    \mathbf{P}^{\mathrm{loc}}_t
    =
    \operatorname{Sinkhorn}
    \left(
        \mathbf{C}^{\mathrm{loc}}_t;
        \boldsymbol{\mu},
        \boldsymbol{\nu},
        \epsilon
    \right).
    \label{eq:local_sinkhorn}
\end{equation}




Then, the resulting transport matrix $\mathbf{P}^{loc}_{t}$ is used to obtain bidirectionally aligned representations through barycentric projection, such as:
\begin{equation}
    \mathbf{V}^{ot}_t
    =\hat{\mathbf{V}}_t
    +\mathrm{Diag}(\boldsymbol{\mu})^{-1}
    \mathbf{P}^{loc}_t\hat{\mathbf{W}},
\end{equation}
\begin{equation}
    \mathbf{W}^{ot}_t
    =\hat{\mathbf{W}}
    +\mathrm{Diag}(\boldsymbol{\nu})^{-1}
    (\mathbf{P}^{loc}_t)^{\top}\hat{\mathbf{V}}_t.
\end{equation}
where $\mathrm{Diag}(\cdot)$ denotes the operation that constructs a diagonal matrix from its input vector.
Here, $\hat{\mathbf{V}}_t$ and $\hat{\mathbf{W}}$ collect normalized visual and word embeddings, respectively. Thus, $\mathbf{V}^{ot}_t\in\mathbb{R}^{M\times D}$ enriches each visual patch with its textual context, while $\mathbf{W}^{ot}_t\in\mathbb{R}^{L'\times D}$ injects the corresponding visual context into each valid word token.

The OT-aligned representations are subsequently fused using a small set of learnable bottleneck tokens. Let $\mathbf{B}\in\mathbb{R}^{N_b\times D}$ denote the bottleneck tokens, where $N_b\ll M,L'$. They first attend to the aligned visual and textual tokens independently as:
\begin{equation}
\small{
    \mathbf{B}_v
    =\mathrm{Attn}(\mathbf{B},\mathbf{V}^{ot}_t,\mathbf{V}^{ot}_t),
    \ \ 
    \mathbf{B}_w
    =\mathrm{Attn}(\mathbf{B},\mathbf{W}^{ot}_t,\mathbf{W}^{ot}_t).}
\end{equation}
where $\mathrm{Attn}(\cdot)$ denotes the multi-head cross-attention operation.
Then, the modality-specific representations are integrated with the
original bottleneck tokens through residual aggregation, followed by
a feed-forward network:
\begin{equation}
    \tilde{\mathbf{B}}
    =\mathrm{FFN}(\mathbf{B}+\mathbf{B}_v+\mathbf{B}_w).
\end{equation}
The updated bottleneck tokens then broadcast the fused information back to the visual tokens:

\begin{equation}
    \tilde{\mathbf{V}}_t = \mathbf{V}_t+g_o(\mathbf{V}^{ot}_t
    +\mathrm{Attn}(\mathbf{V}^{ot}_t,\tilde{\mathbf{B}},\tilde{\mathbf{B}})).
\end{equation}

\subsection{Spatial Multi-Scale Optimal Transport}

To capture spatial structures at multiple resolutions, we construct multi-scale feature representations and align them via spatial-aware optimal transport.
Specifically,
we first reshape the fused visual tokens $\tilde{\mathbf{V}}_t$ into the event feature map $\tilde{\mathbf{X}}_t\in\mathbb{R}^{C_v\times H/p\times W/p}$.
Then, we derive three feature branches from $\tilde{\mathbf{X}}_t$, corresponding to small, middle, and large spatial scales as follows:
\begin{equation}
\small{
    \mathbf{F}_s = \mathrm{Down}(\tilde{\mathbf{X}}_t), \quad
    \mathbf{F}_m = \phi(\tilde{\mathbf{X}}_t), \quad
    \mathbf{F}_l = \mathrm{Up}(\tilde{\mathbf{X}}_t),}
\end{equation}
where $\mathbf{F}_s$ and $\mathbf{F}_l$ are resized to $8 \times 8$ and $32 \times 32$, respectively, and $\phi(\cdot)$ is a $1 \times 1$ projection for the middle branch.
We apply two spatial-aware OT operations to align the small- and large-scale branches with the middle branch. Since both operations share the same procedure, we detail only the small-to-middle alignment. 
Let $\{\mathbf{f}^{s}_i\}_{i=1}^{M_s}$ and $\{\mathbf{f}^{m}_j\}_{j=1}^{M}$ denote the flattened small- and middle-scale tokens.
Their corresponding normalized spatial coordinates are denoted by $\{\mathbf{r}^{s}_{i}\}_{i=1}^{M_s}$ and $\{\mathbf{r}^{m}_{j}\}_{j=1}^{M_m}$.
We define the spatially aware transport cost by jointly considering
the feature discrepancy and spatial distance as:
\begin{equation}
    \mathbf{C}^{s}_{t,ij}
    =
    \lambda_f\|\mathbf{f}^{s}_i-\mathbf{f}^{m}_j\|_2
    +\lambda_p\|\mathbf{r}^{s}_i-\mathbf{r}^{m}_j\|_2,
\end{equation}
where we fix $\lambda_f=0.9$ and $\lambda_p=0.1$ in all experiments. 
Based on this cost matrix, we formulate the small-to-middle cross-scale alignment as an entropically regularized optimal
transport problem and solve it by the same log-domain Sinkhorn iterations as introduced in Appendix~\ref{app:sinkhorn} as:
\begin{equation}
    \mathbf{P}^{s}_t=\mathrm{Sinkhorn}(\mathbf{C}^{s}_t; \boldsymbol{\mu}_s,\boldsymbol{\nu}_m, \varepsilon_s),
\end{equation}
where $\boldsymbol{\mu}_s$ and $\boldsymbol{\nu}_m$ are the uniform source and target marginals, assigning equal transport mass to all tokens at the small and middle scales, respectively. The parameter $\epsilon_s$ is the entropic regularization coefficient, which is set to $0.08$ to obtain a smooth and numerically stable transport plan.
The middle-scale tokens are transported onto the small-scale support and then interpolated back to the middle resolution:
\begin{equation}
    \hat{\mathbf{F}}_s
    =\mathrm{Up}\!\left(
    \mathrm{Reshape}\!\left(
    \mathbf{P}^{s}_t\mathbf{F}^{tok}_m
    \right)\right),
\end{equation}
where $\mathbf{F}^{tok}_m\in\mathbb{R}^{M\times C_v}$ denotes the flattened middle-scale tokens. 

Similarly, for the large-to-middle branch, we first construct the cost matrix $\mathbf{C}^{l}_t$ and then compute the corresponding
transport plan $\mathbf{P}^{l}_t$ following the same procedure. 
The middle-scale tokens are subsequently transported onto the large-scale support, and the resulting features are downsampled to obtain $\hat{\mathbf{F}}_l$.
Finally, the aligned small- and large-scale features are concatenated with the original middle-scale feature to form the multi-scale representation:
\begin{equation}
    \mathbf{X}^{ms}_t =
    \mathrm{Concat}(\hat{\mathbf{F}}_s, \mathbf{F}_m, \hat{\mathbf{F}}_l).
\end{equation}
The resulting feature map $\mathbf{X}^{\mathrm{ms}}_t \in\mathbb{R}^{3C_v\times H/p\times W/p}$
integrates spatial information across multiple resolutions and is subsequently fed into the temporal inter-frame OT module.

\subsection{Temporal Inter-Frame Optimal Transport}

Event streams contain rich temporal dynamics that cannot be fully represented by a single frame. We therefore introduce temporal inter-frame OT to align the spatially enhanced feature maps to a fixed temporal anchor. Let $\mathbf{X}^{ms}_t$ denote the spatial-OT output for frame $t$, and let $\mathbf{F}_t=[\mathbf{f}_{t,1},\ldots,\mathbf{f}_{t,M'}]^{\top}\in\mathbb{R}^{M'\times C'}$ be its flattened token matrix, where the lowercase vector $\mathbf{f}_{t,i}\in\mathbb{R}^{C'}$ denotes the $i$-th spatial token of frame $t$.
For each five-frame input sequence, we select the middle frame (indexed by $a=2$) as the temporal anchor during both training and inference.
Accordingly, $\mathbf{F}_a$ denotes the anchor-frame token matrix and $\mathbf{f}_{a,i}$ its $i$-th token.
For each frame $t\neq a$, we compute the pairwise temporal matching cost between the anchor tokens and the current-frame tokens. 
Both token features are first $\ell_2$-normalized, and the cost is defined as the cosine distance:
\begin{equation}
    \mathbf{C}^{temp}_{t,ij}
    =1-
    \frac{\mathbf{f}_{a,i}^{\top}\mathbf{f}_{t,j}}
    {\|\mathbf{f}_{a,i}\|_2\|\mathbf{f}_{t,j}\|_2}.
\end{equation}
Here, $\mathbf{C}^{temp}_t\in\mathbb{R}^{M'\times M'}$ compares every anchor token $\mathbf{f}_{a,i}$ (row $i$) with every token $\mathbf{f}_{t,j}$ from frame $t$ (column $j$).
Using uniform marginals over the two token sets, we compute the temporal transport plan as
\begin{equation}
    \mathbf{P}^{\mathrm{temp}}_t
    =
    \operatorname{Sinkhorn}
    \left(\mathbf{C}^{\mathrm{temp}}_t; \boldsymbol{\mu}_a,\boldsymbol{\nu}_t, \varepsilon_{\mathrm{temp}}\right),
    \  t\neq a,
    \label{eq:temporal_ot_plan}
\end{equation}
which follows the same log-domain Sinkhorn procedure introduced in Appendix~\ref{app:sinkhorn}.
Here, $\boldsymbol{\mu}_a$ and $\boldsymbol{\nu}_t$ are the uniform marginals that assign equal transport mass to all tokens in the anchor frame and the current frame, respectively. The parameter $\epsilon_{\mathrm{temp}}$ is the entropic regularization coefficient, which is set to $0.08$. We perform 20 Sinkhorn iterations to obtain the
temporal transport plan.

The transport plan is row-normalized and subsequently used to transport the current-frame tokens onto the anchor-frame support:
\begin{equation}
    \widetilde{\mathbf{F}}_t
    = \operatorname{Diag}
    \left(
        \mathbf{P}^{\mathrm{temp}}_t\mathbf{1}
    \right)^{-1}
    \mathbf{P}^{\mathrm{temp}}_t\mathbf{F}_t,
    t \neq a.
\end{equation}
For the anchor itself, we use the identity correspondence, i.e., $\widetilde{\mathbf{F}}_a=\mathbf{F}_a$. This produces a temporally aligned token sequence $\{\widetilde{\mathbf{F}}_t\}_{t=1}^{T}$, all represented on the anchor-frame token support.

To assign different importance to different temporal observations, we first average the tokens of each aligned frame and predict a scalar temporal logit with a lightweight MLP:
\begin{equation}
    \boldsymbol{h}_t
    =\frac{1}{M'}\sum_{i=1}^{M'}\widetilde{\mathbf{f}}_{t,i},
    \ 
    \alpha_t
    =\mathrm{Softmax}_{t=1}^{T}\bigl(\mathrm{MLP}(\boldsymbol{h}_t)\bigr).
\end{equation}
The weighted temporal context is then combined with the anchor representation using a learnable sigmoid gate $\gamma$:
\begin{equation}
    \mathbf{F}^{temp}
    =\mathbf{F}_a+\gamma\left(
    \sum_{t=1}^{T}\alpha_t\widetilde{\mathbf{F}}_t-\mathbf{F}_a
    \right),
    \  \gamma=\sigma(\gamma_0).
\end{equation}
where $\gamma_0$ is a learnable scalar initialized to $-2.0$, and $\gamma=\sigma(\gamma_0)$ adaptively balances the anchor feature and the aggregated temporal context.
The same temporal weights are applied to the CLS tokens when available, yielding a temporally fused global token $\mathbf{c}^{temp}$. The fused token matrix $\mathbf{F}^{temp}$ is reshaped into a feature map $\mathbf{X}^{temp}\in\mathbb{R}^{C'\times H'\times W'}$. The pair $(\mathbf{X}^{temp},\mathbf{c}^{temp})$ is then passed to the standalone SALAD aggregation layer~\cite{mahjoub2023salad}. SALAD performs its learnable cluster assignment and residual aggregation on the temporally fused visual features and returns the final compact place descriptor:
\begin{equation}
    \mathbf{d}=\mathrm{SALAD}(\mathbf{X}^{temp},\mathbf{c}^{temp}).
\end{equation}
The descriptor $\mathbf{d}$ is $\ell_2$-normalized by SALAD and is used directly for place retrieval and the retrieval loss during training.

\subsection{Training Objective}
\label{sec:objective}

We optimize MSTOT-EPR using three complementary objectives.
The retrieval loss $\mathcal{L}_{\mathrm{retr}}$ learns discriminative place descriptors using the Multi-Similarity loss~\cite{wang2019multi}.
The local alignment loss $\mathcal{L}_{\mathrm{loc}}$ minimizes the transport cost between event patches and textual tokens, where gradients are not propagated through the transport plan.
The global alignment loss $\mathcal{L}_{\mathrm{mmd}}$ reduces the distribution discrepancy between the event CLS features and sentence embeddings using multi-kernel MMD.

The complete training objective is
\begin{equation}
\mathcal{L}
=
\mathcal{L}_{\mathrm{retr}}
+\lambda\mathcal{L}_{\mathrm{loc}}
+\mu\mathcal{L}_{\mathrm{mmd}}.
\label{eq:total_loss}
\end{equation}

Detailed definitions of the three losses are provided in
Appendix~\ref{app:training_objectives}.

\section{Experiments} \label{sec::experiments}

\subsection{Dataset and Evaluation Metric}

\noindent\textbf{Datasets.}
We evaluate the proposed method on two event-based VPR benchmarks: \textbf{NYC-Event-VPR}~\cite{pan2025nyc} and \textbf{EPRBench}~\cite{wang2026eprbench}. NYC-Event-VPR provides synchronized RGB images and event streams; we therefore report results on both modalities to assess performance across conventional and event-based observations. EPRBench provides scene-level semantic annotations for event data and includes challenging environmental conditions and substantial appearance variations, making it well suited for evaluating text-assisted event-based place recognition.

\noindent\textbf{Evaluation metrics.}
Following the official evaluation protocols of both datasets, we adopt Recall@$K$ (R@$K$) as the evaluation metric, with $K\in\{1,5,10\}$. For each query, database samples are ranked according to the similarity between their predicted descriptors. A query is considered correctly localized if at least one ground-truth reference appears among the top-$K$ retrieved candidates. R@$K$ is the percentage of correctly localized queries over the entire query set; higher values indicate better retrieval performance.

\subsection{Implementation Details}

We use pretrained DINOv2 ViT-B/14 as the visual backbone and frozen CLIP ViT-L/14 for semantic supervision. Each input consists of five consecutive event frames resized to $224\times224$, with the middle frame ($a=2$) selected as the temporal anchor. Unless otherwise specified, the bottleneck dimension is 256, and all OT modules perform 20 Sinkhorn iterations with an entropic regularization coefficient of 0.08. SALAD uses 64 clusters to produce an 8448-dimensional descriptor. The model is trained for 20 epochs using AdamW with an initial learning rate of $6\times10^{-5}$. Complete implementation details are provided in Appendix~\ref{app:implementation}.


\subsection{Comparison with State-of-the-Art Methods}

Table~\ref{tab:comparison-methods} compares MSTOT-EPR with representative state-of-the-art VPR methods under three evaluation settings. MSTOT-EPR achieves the best performance on both event-based benchmarks, obtaining
65.0\%, 78.9\%, and 80.9\% R@1/R@5/R@10 on NYC-Event-VPR-Event and 95.0\%, 98.5\%, and 98.8\% on EPRBench. These results demonstrate the benefit of jointly modeling event-text semantics, multi-scale spatial structure, and inter-frame temporal correspondence. On the RGB split, MSTOT-EPR remains competitive, although it is not specifically optimized for dense RGB appearance cues. A detailed dataset-wise comparison is provided in Appendix~\ref{app:detailed_sota}. 

\begin{table}[t]
\centering
\caption{Comparison with state-of-the-art visual place recognition methods on NYC-Event-VPR-Event, NYC-Event-VPR-RGB, and EPRBench. Results are reported as Recall@$K$ , where $K\in\{1,5,10\}$. The best result in each column is highlighted in bold.} 
\label{tab:comparison-methods}    
\resizebox{\textwidth}{!}{ 
\begin{tabular}{l|l|ccc|ccc|ccc}
\toprule
\multirow{2}{*}{\textbf{Method}} & \multirow{2}{*}{\textbf{Source}} & \multicolumn{3}{c|}{\textbf{NYC-Event-VPR-Event}} & \multicolumn{3}{c|}{\textbf{NYC-Event-VPR-RGB}} & \multicolumn{3}{c}{\textbf{EPRBench}} \\
& & R@1 & R@5 & R@10 & R@1 & R@5 & R@10 & R@1 & R@5 & R@10 \\
\midrule
01. Deep visual geo-loc~\cite{deepvg} & CVPR'22 & 25.6 & 41.9 & 49.1 & 84.9 & 89.3 & 90.4 & 63.2 & 77.9 & 82.8 \\
02. CosPlace~\cite{cosplace} & CVPR'22 & 32.1 & 48.1 & 55.3 & 85.4 & 88.8 & 89.7 & 73.2 & 88.9 & 92.0 \\
03. MixVPR~\cite{ismail2022mixvpr} & WACV'23 & 45.7 & 63.0 & 68.9 & 88.3 & 91.0 & 91.5 & 90.2 & 93.7 & 95.6  \\
04. EigenPlaces~\cite{eigenplaces} & ICCV'23 & 28.5 & 43.4 & 51.2 & 81.4 & 91.7 & \textbf{93.9} & 81.1 & 86.0 & 89.0 \\
05. R2Former~\cite{xu2023r2former} & CVPR'23 & 61.2 & 75.0 & 78.5 & 86.3 & 88.9 & 89.9 & 80.0 & 94.8 & 96.7 \\
06. CricaVPR~\cite{cricavpr}& CVPR'24 & 30.3 & 48.2 & 55.0 & 88.3 & 92.2 & 92.9 & 92.6 & 95.3 & 96.5 \\
07. SelaVPR~\cite{lu2024towards}& ICLR'24 & 43.3 & 57.6 & 62.9 & 87.8 & 91.3 & 91.7 & 89.9 & 93.6 & 95.2 \\ 
08. EMVP~\cite{qiu2024emvp} & NeurIPS'24 & 45.7 & 64.1 & 70.9 & 85.5 & 90.4 & 91.4 & 88.7 & 93.9 & 95.7 \\
09. SALAD~\cite{mahjoub2023salad}& CVPR'24 & 39.6 & 57.8 & 64.1 & 89.1 & 91.8 & 92.3 & 91.3 & 93.4 & 95.3 \\
10. SuperVLAD~\cite{lu2024supervlad}& NeurIPS'24 & 39.7 & 59.2 & 66.6 & 88.8 & 91.5 & 92.4 & 93.3 & 95.0 & 96.2 \\
11. BoQ~\cite{Ali-bey_2024_CVPR} & CVPR'24 & 38.9 & 56.7 & 63.6 & 83.2 & 88.1 & 89.1 & 93.1 & 94.7 & 95.7 \\
12. ImAge~\cite{ImAge} & NeurIPS'25 &42.4  &59.0  &65.0  &85.7  &90.4  &91.0  & 76.2  & 89.7 & 92.7\\
13. PRGS~\cite{zuo2025prgs} & PR'25 & 43.8 & 62.1 & 69.3 & 82.8 & 88.6 & 89.9 & 87.7  & 93.9 & 96.9\\
14. Pair-VPR~\cite{hausler2025pairvpr} & RA-L'25 & 64.0 & 75.2 & 78.4 & 89.4 & 91.9 & 92.4 & 91.9  & 95.7 & 95.8\\
15. FOL~\cite{wang2025focus} & AAAI'25 & 55.5 & 68.2 & 72.5 & 85.1 & 89.3 & 89.9 & 93.8  & 96.0 & 97.1\\
16. SelaVPR++~\cite{lu2025selavpr++} & TPAMI'25& 37.4 & 56.3 & 62.4 & 66.4 & 76.6 & 78.5 & 90.3 &  94.3 & 96.4 \\
17. UniPR-3D~\cite{deng2026_unipr3d} & ECCV'26 & 44.5 & 59.0 & 63.9 & 69.8 & 78.2 & 80.0 & 91.7 & 94.9 & 96.3 \\
18. SAGE~\cite{chen2026sage} & ICLR'26 & 54.9 & 71.2 & 76.7 & \textbf{89.7} & \textbf{92.5} & 93.0 & 83.9 & 90.0 & 92.5 \\
19. A2GC~\cite{li20252_A2GC} & CVPR'26 & 54.9 & 70.7 & 73.4 & 71.1 & 78.5 & 80.3 & 92.8 & 95.3 & 96.8 \\
20. SG-VPR~\cite{wang2026eprbench} &TCSVT'26 & 57.5 & 74.1 & 78.9 & 86.5 & 90.9 & 92.0 & 94.3 & 96.1 & 97.1 \\
\midrule
21. MSTOT-EPR (Ours) & - &  \textbf{65.0}  & \textbf{78.9} & \textbf{80.9} & 81.5 & 89.7 & 90.7 & \textbf{95.0} & \textbf{98.5} & \textbf{98.8} \\
\bottomrule
\end{tabular}} 
\end{table}

\subsection{Ablation Study}  

\begin{table}[t]
\centering
\caption{Ablation study of the proposed components on the EPRBench
dataset. Spatial optimal transport (Spa-OT), temporal optimal transport
(Temp-OT), and bottleneck fusion (BF) are progressively introduced to
evaluate their individual and complementary contributions.}
\label{tab:ablation_key}
\setlength{\tabcolsep}{8pt}
\renewcommand{\arraystretch}{1.0}
\begin{tabular}{c c c c c c}
\toprule
\multicolumn{3}{c}{\textbf{Key Modules}} &
\multicolumn{3}{c}{\textbf{Evaluation Metrics}} \\
\cmidrule(lr){1-3}\cmidrule(lr){4-6}
\textbf{Spa-OT} &
\textbf{Temp-OT} &
\textbf{BF} &
\textbf{R1 (\%)} $\uparrow$ &
\textbf{R5 (\%)} $\uparrow$ &
\textbf{R10 (\%)} $\uparrow$ \\
\midrule
$\times$ & $\times$ & $\times$ & 91.3 & 93.4 & 95.3 \\
$\checkmark$ & $\times$ & $\times$ & 92.3 & 94.6 & 96.6 \\
$\checkmark$ & $\checkmark$ & $\times$ & 93.9 & 96.9 & 98.0 \\
$\checkmark$ & $\times$ & $\checkmark$ & 93.5 & 95.8 & 96.2 \\
$\checkmark$ & $\checkmark$ & $\checkmark$ & \textbf{95.0} & \textbf{98.5} & \textbf{98.8} \\
\bottomrule
\end{tabular}
\end{table}

\paragraph{Effectiveness of key modules.}
Table~\ref{tab:ablation_key} evaluates the three core components on a SALAD-based baseline. Spatial-OT improves R@1 from 91.3\% to 92.3\%, while adding Temporal-OT or bottleneck fusion further raises it to 93.9\% and 93.5\%, respectively. Combining all three components achieves the best results of 95.0\% R@1, 98.5\% R@5, and 98.8\% R@10, demonstrating their complementary contributions to spatial, temporal, and semantic modeling.

\paragraph{Hyperparameter analysis.}
We further investigate the bottleneck dimension, alignment-loss weights, and temporal anchor-frame selection. The complete results are reported in Appendix~\ref{app:additional_ablation}. A bottleneck dimension of 256, balanced loss weights $(\lambda,\mu)=(0.05,0.05)$, and the middle anchor frame ($t=2$) provide the best overall performance.

\subsection{Visualization and Limitation Analysis}

\begin{figure}[!t]
    \centering
    \includegraphics[
        width=1.0\linewidth
    ]{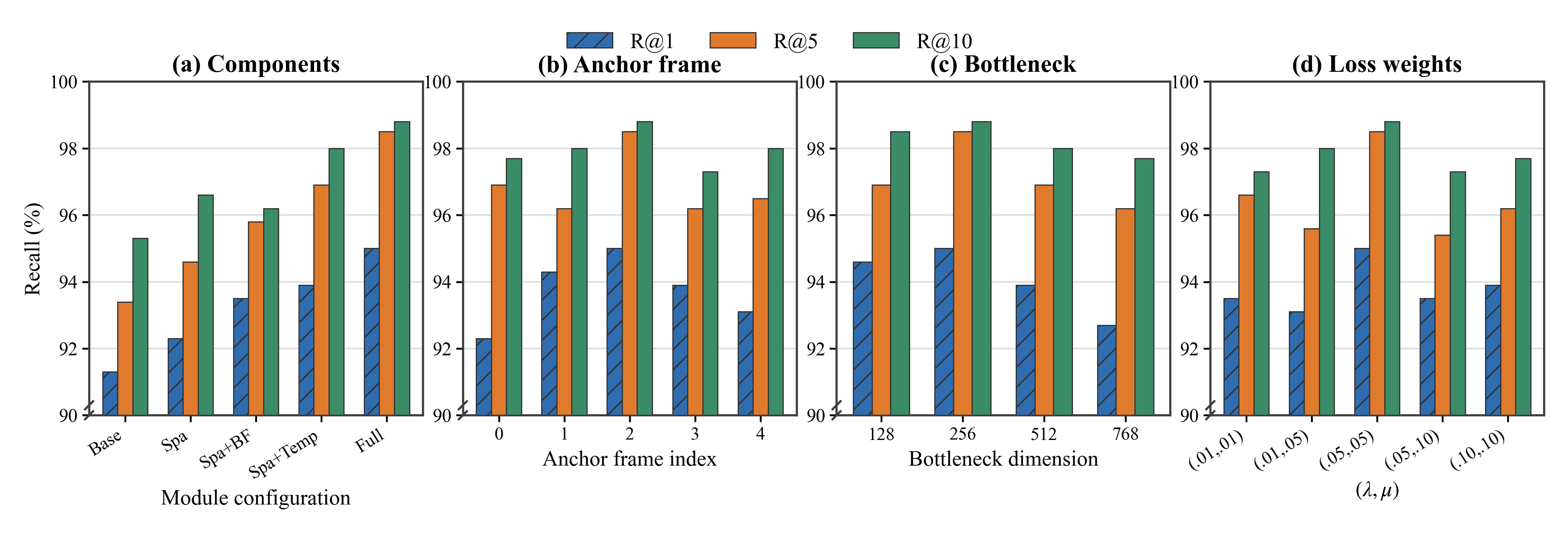}
    \caption{Ablation studies on EPRBench: 
    (a) model components, 
    (b) anchor-frame index,
    (c) bottleneck dimension, and
    (d) alignment-loss weights.}
    \label{fig:ablation_vis}
\end{figure}

Figure~\ref{fig:ablation_vis} summarizes the ablation results of the proposed components and key design choices. Spatial OT improves the SALAD-based baseline from 91.3\% to 92.3\% in R@1, while integrating temporal OT and bottleneck fusion further increases R@1 to 95.0\%. The middle anchor frame ($a=2$), a bottleneck dimension of 256, and balanced alignment-loss weights $(\lambda,\mu)=(0.05,0.05)$ achieve the best overall performance. Additional qualitative retrieval results are provided in Appendix~\ref{app:qualitative}.

While MSTOT-EPR performs well on both event-based benchmarks, it relies on paired text descriptions and is therefore sensitive to annotation noise and
domain shift. Moreover, iterative spatial-temporal OT increases computational cost, while the fixed five-frame window may be suboptimal under rapidly varying event rates. Future work will explore annotation-efficient semantic supervision, sparse OT solvers, and adaptive temporal aggregation.

\section{Conclusion} \label{sec::conclusion}

In this work, we present MSTOT-EPR, an optimal transport-driven framework that bridges high-level text semantics with asynchronous event streams for robust visual place recognition. MSTOT-EPR realizes bidirectional cross-modal alignment via local OT and global MMD supervision with bottleneck fusion, while jointly modeling multi-scale spatial structures and inter-frame dynamics before SALAD-based aggregation. Extensive experiments on NYC-Event-VPR and EPRBench demonstrate our state-of-the-art retrieval accuracy on event data alongside competitive RGB generalization. Ablations confirm the symbiotic contributions of semantic, spatial, and temporal modules. Future work will focus on weakly-supervised semantic distillation, efficient transport solvers, and rate-adaptive temporal aggregation.

\newpage
\subsection*{AI use statement}

In preparing this manuscript, we used generative AI tools (large language models) solely for language refinement and proofreading. These tools were not used to formulate research ideas, design or conduct experiments, implement
code, generate figures, or analyze data. All AI-assisted content was carefully reviewed and revised by the authors, who take full responsibility for the accuracy and integrity of the final manuscript.

\bibliography{iclr2027_conference}
\bibliographystyle{iclr2027_conference}

\appendix

\section{Background and Applications of Optimal Transport}
\label{app:ot_background}

Optimal Transport (OT) provides a principled framework for comparing probability distributions by finding a transport plan that minimizes the cost of moving probability mass between them~\cite{villani2009optimal,peyre2019computational}. Unlike pointwise metrics or global pooling, OT captures soft many-to-many correspondences under marginal constraints. Entropic regularization enables efficient and differentiable optimization through Sinkhorn iterations~\cite{cuturi2013sinkhorn}, making OT practical for modern representation learning.

OT has been applied to domain adaptation~\cite{courty2017optimal}, distribution-aware learning objectives~\cite{frogner2015learning}, style transfer~\cite{kolkin2019style}, and feature matching~\cite{sarlin2020superglue}. In visual place recognition, SALAD~\cite{mahjoub2023salad} employs OT-inspired assignments to aggregate local image features into compact global descriptors. However, existing OT-based VPR methods mainly focus on static RGB images and use OT for spatial matching or feature aggregation.

Event-based VPR additionally requires heterogeneous event-text alignment and the modeling of spatial and temporal event structures. Unlike prior methods that apply OT within a single modality or representation level, our approach introduces OT along three dimensions: event-text token alignment, multi-scale spatial alignment, and inter-frame temporal alignment. It therefore extends OT from static feature aggregation to joint cross-modal, cross-scale, and temporal modeling for text-assisted event-stream VPR.

\section{Log-Domain Sinkhorn Solver}
\label{app:sinkhorn}

All optimal-transport modules in MSTOT-EPR are solved using the same
log-domain Sinkhorn procedure. Given a cost matrix
$\mathbf{C}\in\mathbb{R}^{n\times m}$, source marginal
$\boldsymbol{\mu}\in\mathbb{R}^{n}$, target marginal
$\boldsymbol{\nu}\in\mathbb{R}^{m}$, and entropic regularization
coefficient $\epsilon$, we define $\beta=-1/\epsilon$.

Let $\mathbf{f}^{(k)}\in\mathbb{R}^{n}$ and
$\mathbf{g}^{(k)}\in\mathbb{R}^{m}$ be the log-domain scaling vectors
at iteration $k$. We initialize
$\mathbf{f}^{(0)}=\mathbf{0}$ and $\mathbf{g}^{(0)}=\mathbf{0}$,
and update them as
\begin{align}
f_i^{(k+1)}
&=
\log\mu_i
-
\log\sum_{j=1}^{m}
\exp\left(
    \beta C_{ij}+g_j^{(k)}
\right),                                                   \\
g_j^{(k+1)}
&=
\log\nu_j
-
\log\sum_{i=1}^{n}
\exp\left(
    \beta C_{ij}+f_i^{(k+1)}
\right).
\end{align}

After $K$ iterations, the transport plan is recovered as
\begin{equation}
P_{ij}
=
\exp\left(
    \beta C_{ij}
    + f_i^{(K)}
    + g_j^{(K)}
\right).
\end{equation}

These alternating updates enforce the prescribed row and column
marginals. Performing the computation in the log domain improves
numerical stability. We use this solver for cross-modal, multi-scale
spatial, and inter-frame temporal optimal transport.

\section{Detailed Training Objectives}
\label{app:training_objectives}

\subsection{Retrieval Loss}

Given a mini-batch of $N$ normalized place descriptors
$\{\mathbf{d}_i\}_{i=1}^{N}$, their pairwise similarity is
$s_{ij}=\mathbf{d}_i^\top\mathbf{d}_j$.
For each anchor $i$, the Multi-Similarity miner selects informative
positive and negative sets $\mathcal{S}_i^+$ and $\mathcal{S}_i^-$.
The retrieval loss is
\begin{equation}
\small
\mathcal{L}_{\mathrm{retr}}
=
\frac{1}{N}\sum_{i=1}^{N}
\left[
\frac{1}{\alpha}
\log\left(
1+\sum_{p\in\mathcal{S}_i^+}e^{-\alpha s_{ip}-\tau}
\right)
+
\frac{1}{\beta}
\log\left(
1+\sum_{n\in\mathcal{S}_i^-}e^{\beta s_{in}-\tau}
\right)
\right].
\label{eq:appendix_ms_loss}
\end{equation}

We use $\alpha=1$, $\beta=50$, $\tau=0$, and a mining margin of
$\epsilon_m=0.1$.

\subsection{Local Optimal-Transport Alignment}

Using the cross-modal cost matrix $\mathbf{C}^{\mathrm{loc}}_{i,a}$ and transport plan
$\mathbf{P}^{\mathrm{loc}}_{i,a}$ defined in Sec.~\ref{sec:ot_aligned_fusion}, the loss evaluated on anchor frame $a$ is
\begin{equation}
\mathcal{L}_{\mathrm{loc}}
=
\frac{1}{N}
\sum_{i=1}^{N}
\left\langle
\mathbf{P}^{\mathrm{loc}}_{i,a},
\mathbf{C}^{\mathrm{loc}}_{i,a}
\right\rangle_F.
\label{eq:appendix_local_loss}
\end{equation}
Gradients are not propagated through the transport plan during optimization.

\subsection{Global Distribution Alignment}

Let $\bar{\mathbf{c}}_i$ and $\bar{\mathbf{s}}_i$ denote the CLS feature of the projected
and normalized event and the sentence feature, respectively.
Their distribution discrepancy is minimized using
\begin{equation}
\mathcal{L}_{\mathrm{mmd}}
=
\frac{1}{N^2}\sum_{i,j=1}^{N}
\left[
k(\bar{\mathbf{c}}_i,\bar{\mathbf{c}}_j)
+
k(\bar{\mathbf{s}}_i,\bar{\mathbf{s}}_j)
-
2k(\bar{\mathbf{c}}_i,\bar{\mathbf{s}}_j)
\right].
\label{eq:appendix_mmd_loss}
\end{equation}

Here, $k(\cdot,\cdot)$ is a multi-kernel Gaussian kernel composed of five geometrically scaled bandwidths.

\section{Implementation Details}
\label{app:implementation}

We implement MSTOT-EPR using PyTorch Lightning and train it on a single NVIDIA RTX 3090 GPU. Each input sample consists of five consecutive event frames resized to $224\times224$. The frames are normalized using ImageNet statistics and augmented with RandAugment during training. Each mini-batch contains 20 places, with four event sequences sampled for each place.
We use pretrained DINOv2 ViT-B/14 as the visual backbone and fine-tune only its last four Transformer blocks, while keeping the CLIP ViT-L/14 text encoder frozen. The spatial OT module constructs multi-scale representations using scale factors of 0.5, 1.0, and 2.0. The temporal OT module aligns all five frames to the middle frame ($a=2$). All OT modules perform 20 log-domain Sinkhorn iterations with an entropic regularization coefficient of 0.08. The bottleneck dimension is set to 256, and the alignment-loss weights are set to $(\lambda,\mu)=(0.05,0.05)$. SALAD uses 64 clusters and produces an 8448-dimensional $\ell_2$-normalized place descriptor.
The model is trained using the Multi-Similarity loss and Multi-Similarity Miner. We use AdamW with an initial learning rate of $6\times10^{-5}$, a weight decay of $9.5\times10^{-9}$, and a linear learning-rate scheduler. Training lasts for 20 epochs using mixed-precision computation.

\section{Detailed Comparison with State-of-the-Art Methods}
\label{app:detailed_sota}

\paragraph{NYC-Event-VPR-Event.}
MSTOT-EPR achieves 65.0\%, 78.9\%, and 80.9\% in R@1, R@5, and R@10, respectively. Compared with the best competing result for each metric, the corresponding improvements are 1.0, 3.7, and 2.0 percentage points. The improvements indicate that multi-scale spatial modeling and inter-frame alignment facilitate the extraction of discriminative place
representations from sparse event observations.

\paragraph{EPRBench.}
MSTOT-EPR obtains 95.0\%, 98.5\%, and 98.8\% in R@1, R@5, and R@10, outperforming the best competing results by 0.7, 2.4, and 1.7 percentage points, respectively. The consistent improvements suggest that textual semantics complement sparse event observations, while the
spatial-temporal OT modules preserve structural and temporal consistency
under challenging environmental variations.

\paragraph{NYC-Event-VPR-RGB.}
On the RGB split, MSTOT-EPR achieves 81.5\%, 89.7\%, and 90.7\% in R@1, R@5, and R@10, respectively. Although it does not outperform RGB-specific approaches, the results show that the learned representation retains some generalization to conventional visual inputs.
This performance difference is expected because MSTOT-EPR is specifically designed to address the sparsity and asynchronous dynamics of event data.

\section{Additional Ablation Studies}
\label{app:additional_ablation}

\paragraph{Bottleneck dimension.}
Table~\ref{tab:ablation_dimension} evaluates the bottleneck dimension. The model achieves the best performance with a dimension of 256, reaching 95.0\% R@1, 98.5\% R@5, and 98.8\% R@10. Increasing the dimension to 512 or 768 reduces R@1 to 93.9\% and 92.7\%, respectively, suggesting that an excessively large bottleneck introduces redundant parameters and increases the risk of overfitting.

\begin{table}[t]
    \centering
    \caption{Analysis of bottleneck fusion dimension.}
    \label{tab:ablation_dimension}
    \setlength{\tabcolsep}{3.5mm}
    \begin{tabular}{cccc}
        \toprule
        \textbf{Dimension}
        & \textbf{R@1}
        & \textbf{R@5}
        & \textbf{R@10} \\
        \midrule
        128 & 94.6 & 96.9 & 98.5 \\
        \textbf{256} & \textbf{95.0} & \textbf{98.5} & \textbf{98.8} \\
        512 & 93.9 & 96.9 & 98.0 \\
        768 & 92.7 & 96.2 & 97.7 \\
        \bottomrule
    \end{tabular}
\end{table}

\paragraph{Alignment-loss weights.}
Table~\ref{tab:ablation_loss} studies the weights of the local OT and
global MMD losses. The balanced setting
$(\lambda,\mu)=(0.05,0.05)$ achieves the best overall performance.
Smaller weights provide insufficient cross-modal supervision, whereas
larger weights may overwhelm the primary retrieval objective.

\paragraph{Anchor-frame selection.}
Table~\ref{tab:ablation_anchor} compares different temporal anchors.
The middle frame ($t=2$) performs best because it receives contextual
information from both preceding and subsequent frames.

\begin{table}[!t]
    \centering
    \begin{minipage}[t]{0.45\linewidth}
        \centering
        \caption{Alignment-loss weight analysis.}
        \label{tab:ablation_loss}
        \vspace{3pt}

        \small
        \setlength{\tabcolsep}{1.5mm}
        \renewcommand{\arraystretch}{0.95}
        \begin{tabular}{@{}cccc@{}}
            \toprule
            \textbf{Weight} $(\lambda,\mu)$
            & \textbf{R@1}
            & \textbf{R@5}
            & \textbf{R@10} \\
            \midrule
            $(0.01,0.01)$ & 93.5 & 96.6 & 97.3 \\
            $(0.01,0.05)$ & 93.1 & 95.6 & 98.0 \\
            $\mathbf{(0.05,0.05)}$
            & \textbf{95.0} & \textbf{98.5} & \textbf{98.8} \\
            $(0.05,0.1)$ & 93.5 & 95.4 & 97.3 \\
            $(0.1,0.1)$ & 93.9 & 96.2 & 97.7 \\
            \bottomrule
        \end{tabular}
    \end{minipage}%
    \hspace{0.02\linewidth}%
    \begin{minipage}[t]{0.45\linewidth}
        \centering
        \caption{Anchor-frame selection analysis.}
        \label{tab:ablation_anchor}
        \vspace{3pt}

        \small
        \setlength{\tabcolsep}{1.5mm}
        \renewcommand{\arraystretch}{0.95}
        \begin{tabular}{@{}cccc@{}}
            \toprule
            \textbf{Index} $t$
            & \textbf{R@1}
            & \textbf{R@5}
            & \textbf{R@10} \\
            \midrule
            0 & 92.3 & 96.9 & 97.7 \\
            1 & 94.3 & 96.2 & 98.0 \\
            \textbf{2}
            & \textbf{95.0} & \textbf{98.5} & \textbf{98.8} \\
            3 & 93.9 & 96.2 & 97.3 \\
            4 & 93.1 & 96.5 & 98.0 \\
            \bottomrule
        \end{tabular}
    \end{minipage}
\end{table}

\section{Additional Qualitative Results}
\label{app:qualitative}

\begin{figure}[t]
\centering
\includegraphics[width=1\linewidth]{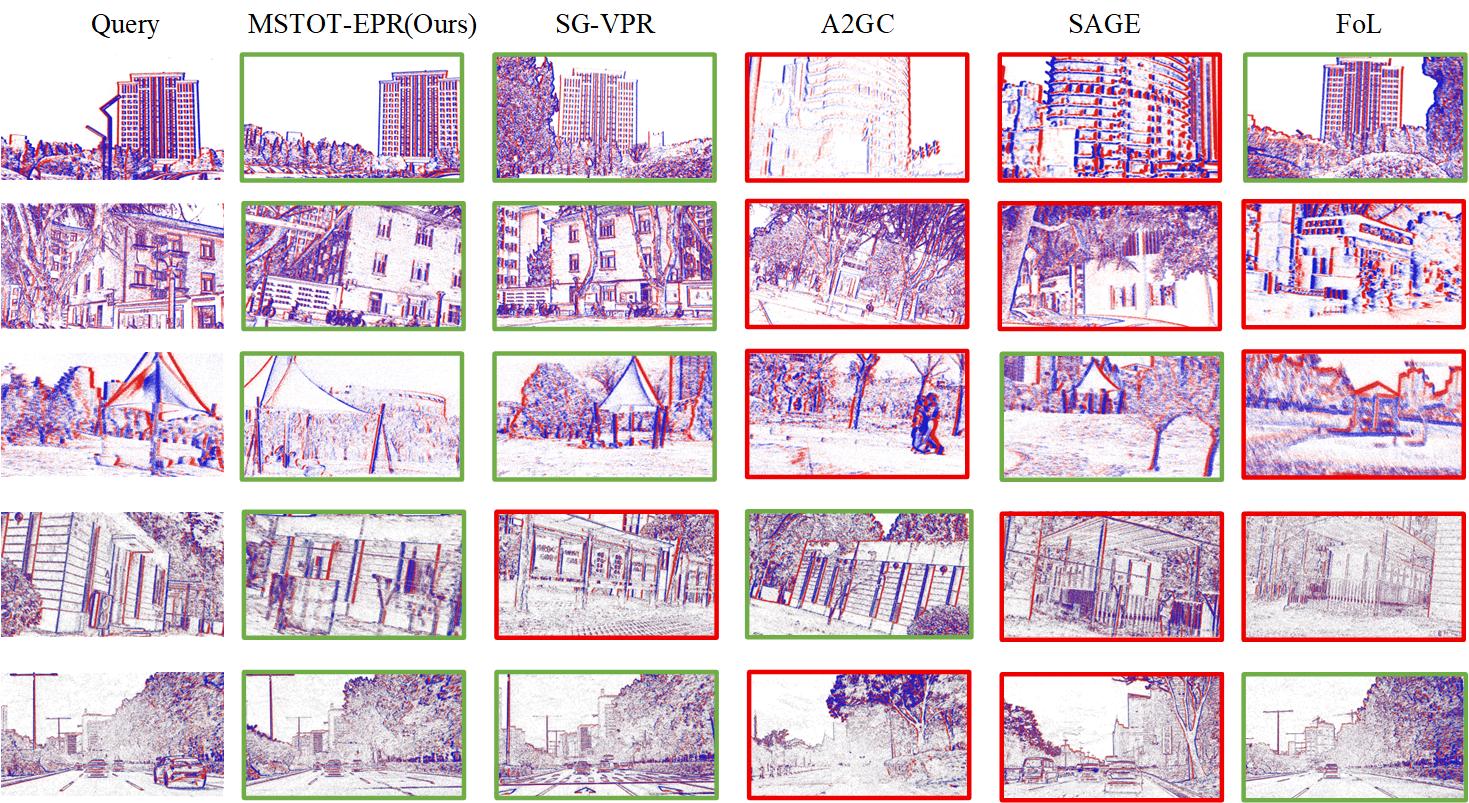}
\caption{Qualitative top-1 retrieval comparison on EPRBench. Each row shows an event-stream query and the corresponding top-1 retrieved reference image from MSTOT-EPR (ours) and competing methods. Green and red borders denote correct and incorrect retrievals, respectively. Our method retrieves the correct places more consistently under viewpoint changes, sparse event observations, and challenging scene appearances.} 
\label{fig:retrieval_vis}
\end{figure}

Figure~\ref{fig:retrieval_vis} presents qualitative top-1 retrieval
comparisons on EPRBench. The selected queries cover diverse and challenging
conditions, including high-rise buildings with repetitive facade patterns,
oblique observations of similar architectural structures, vegetation-dominated
scenes, and road environments containing limited structural details. In these
cases, event observations mainly preserve intensity-change contours while
providing limited texture and appearance information. Consequently, different
places may exhibit similar event patterns, whereas observations of the same
place can vary considerably under changes in viewpoint, motion, and event
density. As shown in Figure~\ref{fig:retrieval_vis}, competing methods correctly retrieve several relatively distinctive scenes but produce inconsistent results when local structures are repetitive or only partially observed. In contrast, MSTOT-EPR returns the correct reference for all the presented queries, indicating more stable retrieval behavior across different scene types.

The qualitative results are also consistent with the design of MSTOT-EPR.
First, the event-text alignment module establishes fine-grained correspondences
between event patches and textual tokens, providing complementary scene-level
semantics when event appearance alone is ambiguous. Second, the multi-scale
spatial OT module aligns event representations using both feature similarity
and spatial position, allowing local contours and global layouts to be compared
across different resolutions. This is particularly relevant to scenes involving
repetitive facades, scale changes, or partial structural observations. Third,
the inter-frame temporal OT module transports neighboring frame tokens onto
a shared anchor support and adaptively aggregates the aligned features. It
therefore exploits complementary observations distributed across consecutive
event frames instead of relying on a single sparse representation. The improved
retrieval consistency shown in these examples qualitatively supports the joint
use of semantic guidance, cross-scale spatial correspondence, and inter-frame
temporal alignment for event-based place recognition.

\end{document}